\PassOptionsToPackage{prologue,table}{xcolor}

\documentclass[sigconf]{aamas}

\usepackage{balance} 
\usepackage{multirow}
\usepackage{booktabs}       

\makeatletter
\gdef\@copyrightpermission{
  \begin{minipage}{0.2\columnwidth}
   \href{https://creativecommons.org/licenses/by/4.0/}{\includegraphics[width=0.90\textwidth]{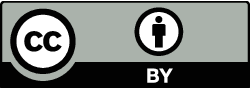}}
  \end{minipage}\hfill
  \begin{minipage}{0.8\columnwidth}
   \href{https://creativecommons.org/licenses/by/4.0/}{This work is licensed under a Creative Commons Attribution International 4.0 License.}
  \end{minipage}
  \vspace{5pt}
}
\makeatother

\setcopyright{ifaamas}
\acmConference[AAMAS '25]{Proc.\@ of the 24th International Conference
on Autonomous Agents and Multiagent Systems (AAMAS 2025)}{May 19 -- 23, 2025}
{Detroit, Michigan, USA}{Y.~Vorobeychik, S.~Das, A.~Nowé  (eds.)}
\copyrightyear{2025}
\acmYear{2025}
\acmDOI{}
\acmPrice{}
\acmISBN{}

\acmSubmissionID{<<OpenReview submission id>>}

\title[AAMAS-2025 Formatting Instructions]{Enhancing Robot Navigation Policies with Task-Specific Uncertainty Managements}
\subtitle{Extended Abstract}

\author{Gokul Puthumanaillam}
\affiliation{
  \institution{University of Illinois Urbana-Champaign}
  \city{Urbana, IL}
  \country{USA}}
\email{gokulp2@illinois.edu}

\author{Paulo Padrao}
\affiliation{
  \institution{Florida International University}
  \city{Miami, FL}
  \country{USA}}
\email{plope113@fiu.edu}

\author{Jose Fuentes}
\affiliation{
  \institution{Florida International University}
  \city{Miami, FL}
  \country{USA}}
\email{jfuen099@fiu.edu}

\author{Leonardo Bobadilla}
\affiliation{
  \institution{Florida International University}
  \city{Miami, FL}
  \country{USA}}
\email{bobadilla@cs.fiu.edu}

\author{Melkior Ornik}
\affiliation{
  \institution{University of Illinois Urbana-Champaign}
  \city{Urbana, IL}
  \country{USA}}
\email{mornik@illinois.edu}

\begin{abstract}
Robots navigating complex environments must manage uncertainty from sensor noise, environmental changes, and incomplete information, with different tasks requiring varying levels of precision in different areas. For example, precise localization may be crucial near obstacles but less critical in open spaces. We present GUIDE (\underline{G}eneralized \underline{U}ncertainty \underline{I}ntegration for \underline{D}ecision-Making and \underline{E}xecution), a framework that integrates these task-specific requirements into navigation policies via Task-Specific Uncertainty Maps (TSUMs). By assigning acceptable uncertainty levels to different locations, TSUMs enable robots to adapt uncertainty management based on context. When combined with reinforcement learning, GUIDE learns policies that balance task completion and uncertainty management without extensive reward engineering. Real-world tests show significant performance gains over methods lacking task-specific uncertainty awareness.

\end{abstract}


\newcommand{\BibTeX}{\rm B\kern-.05em{\sc i\kern-.025em b}\kern-.08em\TeX}

\begin{document}


\pagestyle{fancy}
\fancyhead{}


\maketitle 


\section{Introduction}
Robots encounter varying degrees of uncertainty in sensor measurements, motion models, and environmental conditions. Critically, not all tasks require uniform levels of certainty: for instance, a robot navigating tight corridors must localize precisely, while crossing open spaces may tolerate higher uncertainty. This \emph{context-dependent} nature of uncertainty requirements motivates a need for policies that can selectively reduce or accept uncertainty based on task-specific demands.

Existing methods either aim to minimize uncertainty everywhere \cite{1a,1k,1g} or enforce fixed thresholds \cite{1h,1j}, often leading to inefficiency when uncertainty requirements vary within the same task. Some approaches attempt a uniform trade-off between task performance and uncertainty management \cite{1l,1m, puthumanaillam2024tab}, but they still treat uncertainty needs as globally homogeneous. Moreover, reward engineering or manual tuning \cite{1n,1o} is frequently required to capture task constraints.
In parallel, extensive work in probabilistic robotics \cite{1p,1q,1t,1u} addresses uncertainty but does so uniformly across the workspace. Reinforcement learning (RL) has proven effective for navigation \cite{1v,2f, puthumanaillam2024weathering} but seldom integrates task-specific uncertainty considerations \cite{2g,2h}. Techniques that penalize high-uncertainty actions \cite{2i,2k,2l,2m} or use Bayesian RL \cite{2n,2o,2v,2w} still adopt a one-size-fits-all approach. Risk-aware planning and risk-sensitive RL \cite{3m,3n,3o,3p,3t} similarly rely on uniform thresholds. Although language-conditioned methods \cite{3u,3v,4a} broaden a robot’s task repertoire, they lack a direct mechanism to represent and leverage region-specific uncertainty limits.

We address this gap by introducing \emph{Task-Specific Uncertainty Maps (TSUMs)} that encode allowable uncertainty levels across different regions of the environment for a given task. TSUM captures the \emph{context-dependent value of certainty}, enabling robots to focus on precision only where it is crucial. We present a policy-conditioning framework, \emph{GUIDE} (\underline{G}eneralized \underline{U}ncertainty \underline{I}ntegration for \underline{D}ecis-ion-Making and \underline{E}xecution), which integrates TSUMs into navigation policies. We adapt Soft Actor-Critic (SAC) to \emph{GUIDEd SAC}, balancing task objectives and uncertainty reduction without ad-hoc reward engineering.

\section{Methodology}
Consider a robot operating in a continuous state space \(S\) with a continuous action set \(A\). 
 \begin{figure*}[htbp]
  \centering
  \includegraphics[trim=0pt 50pt 0pt 30pt, clip, width=0.9\textwidth]{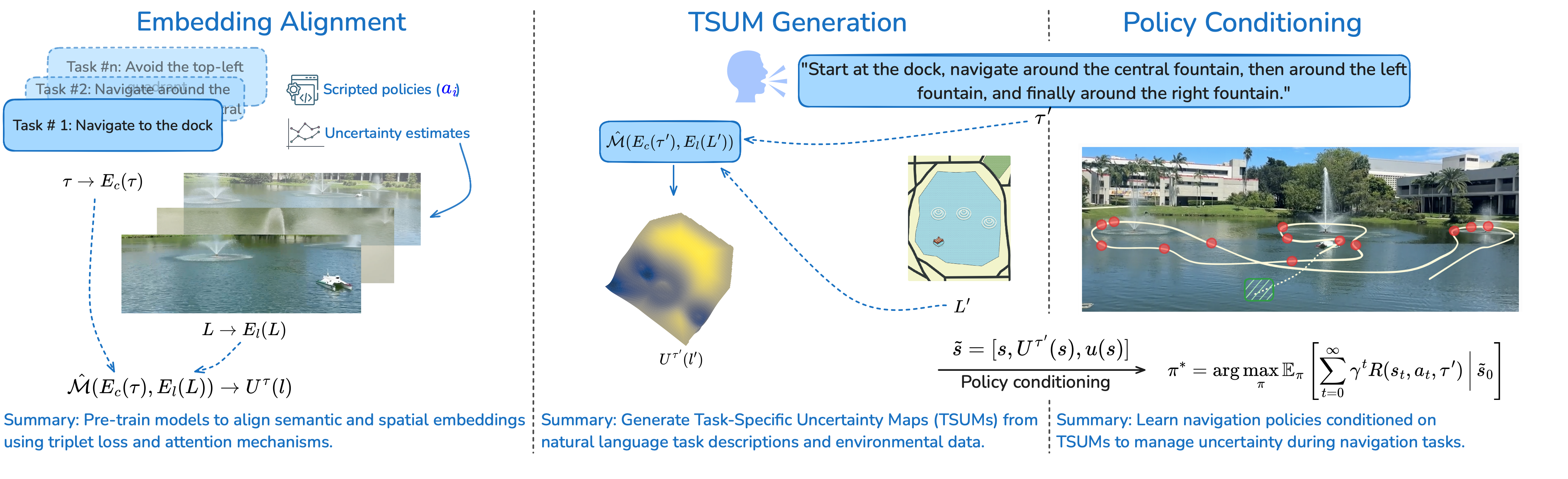}
  \caption{During pretraining, semantic and spatial embeddings are aligned via triplet loss and attention. At deployment, TSUMs derived from task descriptions and environment data condition the navigation policy for task-aware uncertainty management.}
  \label{fig:architecture}
\end{figure*}
Given a navigation task \(\tau\) specified in natural language, the objective is to learn a policy \(\pi(a \mid s)\) that jointly satisfies task objectives and task-specific uncertainty requirements. 

\textit{Task-Specific Uncertainty Maps (TSUM):}
A TSUM is defined as \(U^\tau(l)\), a scalar representing the acceptable uncertainty at location \(l\). Formally,
$U^\tau(l) = w_\Phi\,\Phi^\tau(l) + w_{\mathcal{C}}\,\mathcal{C}^\tau(l) + w_{\mathcal{E}}\,\mathcal{E}(l)$,
where \(\Phi^\tau(l)\) captures the \emph{relevance} of \(l\) for the task, \(\mathcal{C}^\tau(l)\) encodes \emph{constraints} such as safety and legal restrictions, and \(\mathcal{E}(l)\) reflects \emph{environmental factors}. Task semantics are extracted from the natural language specification \(\tau\) using a RoBERTa-based parser \cite{g1}, which identifies subtasks and constraints. Each location \(l\) is mapped to a spatial embedding via a neural network that processes coordinate and environment features. Alignment between these spatial embeddings and the text-derived subtasks is enforced through a triplet loss that brings related concepts closer in embedding space while separating unrelated pairs. An attention mechanism \cite{g3} weights the most relevant subtasks or constraints for each location, producing a single scalar \(U^\tau(l)\) indicating how critical it is to maintain low uncertainty at \(l\). Figure \ref{fig:architecture} (center) represents an example TSUM. 

\textit{Policy Conditioning: }
Once the TSUM is generated, the robot’s state is augmented with both the TSUM value and the robot’s current uncertainty. Concretely, if \(s\) denotes the original state, then the augmented state is \(\tilde{s} = [s,\, U^\tau(s),\, u(s)]\), where \(u(s)\) represents the robot’s current state-estimation uncertainty at \(s\). The policy \(\pi(a \mid \tilde{s})\) thus explicitly observes how precise the localization at \(s\) needs to be, enabling actions that selectively reduce uncertainty in regions with stricter tolerances.

To implement this, we adopt a variant of SAC \cite{3l}, referred to as GUIDEd SAC (G-SAC). Standard SAC updates the policy and Q-value networks to maximize expected reward while promoting exploratory behavior via an entropy term. In G-SAC, the Q-function and policy networks receive \(\tilde{s}\) as input, allowing them to incorporate the TSUM-derived acceptable uncertainty as part of the state. By conditioning on \(U^\tau(s)\), the learned policy mitigates uncertainty precisely in those regions where the task demands high localization accuracy, while avoiding unnecessary effort in areas where uncertainty can be higher without affecting task success--eliminating the need for extensive reward tuning or ad hoc penalty terms, as the TSUM itself encodes spatially varying uncertainty requirements.

\section{Experiments and Results}
We evaluate GUIDE on real-world navigation tasks using an autonomous surface vehicle (ASV) operating in a lake with obstacles and environmental disturbances. By default, the ASV employs noisy state estimation (low cost, high uncertainty) but can temporarily request precise GPS data at an added penalty \cite{puthumanaillam2024comtraq, w5}.

\textit{Baselines and Ablations.} We compare GUIDE (implemented as G-SAC) against: \textbf{SAC} (no TSUMs), \textbf{SAC-P} (penalized uncertainty), \textbf{B-SAC} (bootstrapped uncertainty), \textbf{CVaR} (risk-sensitive RL), \textbf{RAA} (risk-aware planning), and \textbf{HEU} (handcrafted policy that switches to GPS near obstacles). All methods share the same environment observations and cost structure, providing a fair evaluation of task-specific uncertainty handling.

\textit{Performance Comparison.} Table~\ref{tab:comparison-condensed} summarizes Task Completion Rate (TCR) and reward (R) for four representative tasks: \emph{Goal Reaching (GR)}, \emph{Avoid (AV)}, \emph{Perimeter (PT)}, and \emph{Multi-Goal (MG)}. G-SAC achieves the highest TCR and reward in all tasks, demonstrating safer navigation, fewer collisions, and more cost-effective use of precise GPS. Figure~\ref{fig:architecture} (right) illustrates how G-SAC requests exact localization only in areas where the Task-Specific Uncertainty Map (TSUM) dictates tighter uncertainty requirements.
\begin{table}[h]
\centering
\small
\renewcommand{\arraystretch}{1.0}
\setlength{\tabcolsep}{4pt} 
\begin{tabular}{l|cc|cc|cc|cc}
\toprule
 & \multicolumn{2}{c|}{\textbf{GR}} & \multicolumn{2}{c|}{\textbf{AV}} & \multicolumn{2}{c|}{\textbf{PT}} & \multicolumn{2}{c}{\textbf{MG}} \\
\textbf{Method} & \textbf{TCR} & \textbf{R} & \textbf{TCR} & \textbf{R} & \textbf{TCR} & \textbf{R} & \textbf{TCR} & \textbf{R} \\
\midrule
SAC & 68.9 & 144.1 & 71.3 & 177.8 & 44.3 & 84.4 & 31.3 & 124.4 \\
SAC-P & 84.3 & 241.8 & 83.2 & 199.2 & 51.6 & 132.8 & 42.9 & 135.2 \\
B-SAC & 74.2 & 189.2 & 79.6 & 277.6 & 56.3 & 170.4 & 37.7 & 122.4 \\
CVaR & 66.8 & 134.6 & 78.4 & 220.4 & 41.6 & 32.8 & 30.9 & 129.2 \\
RAA & 35.3 & 26.9 & 51.3 & 107.8 & 39.8 & 111.2 & 19.5 & 100.4 \\
HEU & 71.3 & 176.1 & 62.4 & 174.4 & 49.6 & 146.8 & 42.1 & 155.2 \\
\rowcolor{gray!20}
\textbf{G-SAC} & \textbf{92.0} & \textbf{410.4} & \textbf{88.7} & \textbf{482.2} & \textbf{83.7} & \textbf{594.6} & \textbf{81.7} & \textbf{511.4} \\
\bottomrule
\end{tabular}
\caption{Task Completion Rate (TCR) and average reward (R) for different tasks. G-SAC outperforms all baselines.}
\label{tab:comparison-condensed}
\vspace{-6pt}
\end{table}



\begin{acks}
This work is supported in part by NSF grants IIS-2024733 and IIS-2331908, the ONR grants N00014-23-1-2789, N00014-23-1-2651, and N00014-23-1-2505, the DHS grant 23STSLA00016-01-00, the DoD grant 78170-RT-REP, and the Florida Department of Environmental Protection grant INV31.
\end{acks}



\bibliographystyle{ACM-Reference-Format} 
\bibliography{sample}


\begin{thebibliography}{40}


\ifx \showCODEN    \undefined \def \showCODEN     #1{\unskip}     \fi
\ifx \showDOI      \undefined \def \showDOI       #1{#1}\fi
\ifx \showISBNx    \undefined \def \showISBNx     #1{\unskip}     \fi
\ifx \showISBNxiii \undefined \def \showISBNxiii  #1{\unskip}     \fi
\ifx \showISSN     \undefined \def \showISSN      #1{\unskip}     \fi
\ifx \showLCCN     \undefined \def \showLCCN      #1{\unskip}     \fi
\ifx \shownote     \undefined \def \shownote      #1{#1}          \fi
\ifx \showarticletitle \undefined \def \showarticletitle #1{#1}   \fi
\ifx \showURL      \undefined \def \showURL       {\relax}        \fi
\providecommand\bibfield[2]{#2}
\providecommand\bibinfo[2]{#2}
\providecommand\natexlab[1]{#1}
\providecommand\showeprint[2][]{arXiv:#2}

\bibitem[\protect\citeauthoryear{Alali and Imani}{Alali and Imani}{2024}]%
        {2o}
\bibfield{author}{\bibinfo{person}{Mohammad Alali} {and} \bibinfo{person}{Mahdi Imani}.} \bibinfo{year}{2024}\natexlab{}.
\newblock \showarticletitle{Bayesian reinforcement learning for navigation planning in unknown environments}.
\newblock \bibinfo{journal}{\emph{Frontiers in Artificial Intelligence}} (\bibinfo{year}{2024}).
\newblock


\bibitem[\protect\citeauthoryear{Budd, Duckworth, Hawes, and Lacerda}{Budd et~al\mbox{.}}{2023}]%
        {2n}
\bibfield{author}{\bibinfo{person}{Matthew Budd}, \bibinfo{person}{Paul Duckworth}, \bibinfo{person}{Nick Hawes}, {and} \bibinfo{person}{Bruno Lacerda}.} \bibinfo{year}{2023}\natexlab{}.
\newblock \showarticletitle{Bayesian reinforcement learning for single-episode missions in partially unknown environments}. In \bibinfo{booktitle}{\emph{Conference on Robot Learning}}.
\newblock


\bibitem[\protect\citeauthoryear{Chang, Gervet, Khanna, Yenamandra, Shah, Min, Shah, Paxton, Gupta, Batra, et~al\mbox{.}}{Chang et~al\mbox{.}}{2023}]%
        {4a}
\bibfield{author}{\bibinfo{person}{Matthew Chang}, \bibinfo{person}{Theophile Gervet}, \bibinfo{person}{Mukul Khanna}, \bibinfo{person}{Sriram Yenamandra}, \bibinfo{person}{Dhruv Shah}, \bibinfo{person}{So~Yeon Min}, \bibinfo{person}{Kavit Shah}, \bibinfo{person}{Chris Paxton}, \bibinfo{person}{Saurabh Gupta}, \bibinfo{person}{Dhruv Batra}, {et~al\mbox{.}}} \bibinfo{year}{2023}\natexlab{}.
\newblock \showarticletitle{{GOAT: G}o to any thing}.
\newblock \bibinfo{journal}{\emph{arXiv preprint arXiv:2311.06430}} (\bibinfo{year}{2023}).
\newblock


\bibitem[\protect\citeauthoryear{Chen, Gupta, and Gupta}{Chen et~al\mbox{.}}{2019}]%
        {1o}
\bibfield{author}{\bibinfo{person}{Tao Chen}, \bibinfo{person}{Saurabh Gupta}, {and} \bibinfo{person}{Abhinav Gupta}.} \bibinfo{year}{2019}\natexlab{}.
\newblock \showarticletitle{Learning exploration policies for navigation}.
\newblock \bibinfo{journal}{\emph{arXiv preprint arXiv:1903.01959}} (\bibinfo{year}{2019}).
\newblock


\bibitem[\protect\citeauthoryear{Cook and Zhang}{Cook and Zhang}{2020}]%
        {1h}
\bibfield{author}{\bibinfo{person}{Gerald Cook} {and} \bibinfo{person}{Feitian Zhang}.} \bibinfo{year}{2020}\natexlab{}.
\newblock \bibinfo{booktitle}{\emph{Mobile robots: {N}avigation, control and sensing, surface robots and AUVs}}.
\newblock \bibinfo{publisher}{John Wiley \& Sons}.
\newblock


\bibitem[\protect\citeauthoryear{Curtis, Matheos, Gothoskar, Mansinghka, Tenenbaum, Lozano-P{\'e}rez, and Kaelbling}{Curtis et~al\mbox{.}}{2024}]%
        {2g}
\bibfield{author}{\bibinfo{person}{Aidan Curtis}, \bibinfo{person}{George Matheos}, \bibinfo{person}{Nishad Gothoskar}, \bibinfo{person}{Vikash Mansinghka}, \bibinfo{person}{Joshua Tenenbaum}, \bibinfo{person}{Tom{\'a}s Lozano-P{\'e}rez}, {and} \bibinfo{person}{Leslie~Pack Kaelbling}.} \bibinfo{year}{2024}\natexlab{}.
\newblock \showarticletitle{Partially Observable Task and Motion Planning with Uncertainty and Risk Awareness}.
\newblock \bibinfo{journal}{\emph{arXiv preprint arXiv:2403.10454}} (\bibinfo{year}{2024}).
\newblock


\bibitem[\protect\citeauthoryear{Fox, Thrun, Burgard, and Dellaert}{Fox et~al\mbox{.}}{2001}]%
        {1u}
\bibfield{author}{\bibinfo{person}{Dieter Fox}, \bibinfo{person}{Sebastian Thrun}, \bibinfo{person}{Wolfram Burgard}, {and} \bibinfo{person}{Frank Dellaert}.} \bibinfo{year}{2001}\natexlab{}.
\newblock \showarticletitle{Particle filters for mobile robot localization}.
\newblock In \bibinfo{booktitle}{\emph{Sequential Monte Carlo methods in practice}}.
\newblock


\bibitem[\protect\citeauthoryear{Gonzalez-Garcia and Casta{\~n}eda}{Gonzalez-Garcia and Casta{\~n}eda}{2021}]%
        {1k}
\bibfield{author}{\bibinfo{person}{Alejandro Gonzalez-Garcia} {and} \bibinfo{person}{Herman Casta{\~n}eda}.} \bibinfo{year}{2021}\natexlab{}.
\newblock \showarticletitle{Guidance and control based on adaptive sliding mode strategy for a USV subject to uncertainties}.
\newblock \bibinfo{journal}{\emph{IEEE Journal of Oceanic Engineering}} (\bibinfo{year}{2021}).
\newblock


\bibitem[\protect\citeauthoryear{Gul, Rahiman, and Nazli~Alhady}{Gul et~al\mbox{.}}{2019}]%
        {1g}
\bibfield{author}{\bibinfo{person}{Faiza Gul}, \bibinfo{person}{Wan Rahiman}, {and} \bibinfo{person}{Syed~Sahal Nazli~Alhady}.} \bibinfo{year}{2019}\natexlab{}.
\newblock \showarticletitle{A comprehensive study for robot navigation techniques}.
\newblock \bibinfo{journal}{\emph{Cogent Engineering}} (\bibinfo{year}{2019}).
\newblock


\bibitem[\protect\citeauthoryear{Guo, Zhang, Du, Zheng, and Cao}{Guo et~al\mbox{.}}{2021}]%
        {2k}
\bibfield{author}{\bibinfo{person}{Siyu Guo}, \bibinfo{person}{Xiuguo Zhang}, \bibinfo{person}{Yiquan Du}, \bibinfo{person}{Yisong Zheng}, {and} \bibinfo{person}{Zhiying Cao}.} \bibinfo{year}{2021}\natexlab{}.
\newblock \showarticletitle{Path planning of coastal ships based on optimized {DQN} reward function}.
\newblock \bibinfo{journal}{\emph{Journal of Marine Science and Engineering}} (\bibinfo{year}{2021}).
\newblock


\bibitem[\protect\citeauthoryear{Gustafsson, Gunnarsson, Bergman, Forssell, Jansson, Karlsson, and Nordlund}{Gustafsson et~al\mbox{.}}{2002}]%
        {1p}
\bibfield{author}{\bibinfo{person}{Fredrik Gustafsson}, \bibinfo{person}{Fredrik Gunnarsson}, \bibinfo{person}{Niclas Bergman}, \bibinfo{person}{Urban Forssell}, \bibinfo{person}{Jonas Jansson}, \bibinfo{person}{Rickard Karlsson}, {and} \bibinfo{person}{P-J Nordlund}.} \bibinfo{year}{2002}\natexlab{}.
\newblock \showarticletitle{Particle filters for positioning, navigation, and tracking}.
\newblock \bibinfo{journal}{\emph{IEEE Transactions on signal processing}} (\bibinfo{year}{2002}).
\newblock


\bibitem[\protect\citeauthoryear{Haarnoja, Zhou, Abbeel, and Levine}{Haarnoja et~al\mbox{.}}{2018}]%
        {3l}
\bibfield{author}{\bibinfo{person}{Tuomas Haarnoja}, \bibinfo{person}{Aurick Zhou}, \bibinfo{person}{Pieter Abbeel}, {and} \bibinfo{person}{Sergey Levine}.} \bibinfo{year}{2018}\natexlab{}.
\newblock \showarticletitle{Soft actor-critic: Off-policy maximum entropy deep reinforcement learning with a stochastic actor}. In \bibinfo{booktitle}{\emph{International Conference on Machine Learning}}.
\newblock


\bibitem[\protect\citeauthoryear{Hakobyan, Kim, and Yang}{Hakobyan et~al\mbox{.}}{2019}]%
        {3n}
\bibfield{author}{\bibinfo{person}{Astghik Hakobyan}, \bibinfo{person}{Gyeong~Chan Kim}, {and} \bibinfo{person}{Insoon Yang}.} \bibinfo{year}{2019}\natexlab{}.
\newblock \showarticletitle{Risk-aware motion planning and control using CVaR-constrained optimization}.
\newblock \bibinfo{journal}{\emph{IEEE Robotics and Automation letters}} (\bibinfo{year}{2019}).
\newblock


\bibitem[\protect\citeauthoryear{Ko and Fox}{Ko and Fox}{2009}]%
        {1q}
\bibfield{author}{\bibinfo{person}{Jonathan Ko} {and} \bibinfo{person}{Dieter Fox}.} \bibinfo{year}{2009}\natexlab{}.
\newblock \showarticletitle{{GP-BayesFilters}: {B}ayesian filtering using Gaussian process prediction and observation models}.
\newblock \bibinfo{journal}{\emph{Autonomous Robots}} (\bibinfo{year}{2009}).
\newblock


\bibitem[\protect\citeauthoryear{Kumar, Fu, Soh, Tucker, and Levine}{Kumar et~al\mbox{.}}{2019}]%
        {2v}
\bibfield{author}{\bibinfo{person}{Aviral Kumar}, \bibinfo{person}{Justin Fu}, \bibinfo{person}{Matthew Soh}, \bibinfo{person}{George Tucker}, {and} \bibinfo{person}{Sergey Levine}.} \bibinfo{year}{2019}\natexlab{}.
\newblock \showarticletitle{Stabilizing off-policy {Q}-{L}earning via bootstrapping error reduction}.
\newblock \bibinfo{journal}{\emph{Advances in Neural Information Processing Systems}} (\bibinfo{year}{2019}).
\newblock


\bibitem[\protect\citeauthoryear{Li, Yang, Li, Qu, Lyu, and Li}{Li et~al\mbox{.}}{2022}]%
        {3m}
\bibfield{author}{\bibinfo{person}{Guofa Li}, \bibinfo{person}{Yifan Yang}, \bibinfo{person}{Shen Li}, \bibinfo{person}{Xingda Qu}, \bibinfo{person}{Nengchao Lyu}, {and} \bibinfo{person}{Shengbo~Eben Li}.} \bibinfo{year}{2022}\natexlab{}.
\newblock \showarticletitle{Decision making of autonomous vehicles in lane change scenarios: Deep reinforcement learning approaches with risk awareness}.
\newblock \bibinfo{journal}{\emph{Transportation research part C: emerging technologies}} (\bibinfo{year}{2022}).
\newblock


\bibitem[\protect\citeauthoryear{Li, Ma, Cao, Luo, Wang, and Chen}{Li et~al\mbox{.}}{2024}]%
        {2m}
\bibfield{author}{\bibinfo{person}{Yueming Li}, \bibinfo{person}{Mingquan Ma}, \bibinfo{person}{Jian Cao}, \bibinfo{person}{Guobin Luo}, \bibinfo{person}{Depeng Wang}, {and} \bibinfo{person}{Weiqiang Chen}.} \bibinfo{year}{2024}\natexlab{}.
\newblock \showarticletitle{A Method for Multi-AUV Cooperative Area Search in Unknown Environment Based on Reinforcement Learning}.
\newblock \bibinfo{journal}{\emph{Journal of Marine Science and Engineering}} (\bibinfo{year}{2024}).
\newblock


\bibitem[\protect\citeauthoryear{Liu}{Liu}{2019}]%
        {g1}
\bibfield{author}{\bibinfo{person}{Yinhan Liu}.} \bibinfo{year}{2019}\natexlab{}.
\newblock \showarticletitle{Roberta: {A} robustly optimized {BERT} pretraining approach}.
\newblock \bibinfo{journal}{\emph{arXiv preprint arXiv:1907.11692}} (\bibinfo{year}{2019}).
\newblock


\bibitem[\protect\citeauthoryear{Morad, Kortvelesy, Bettini, Liwicki, and Prorok}{Morad et~al\mbox{.}}{2023}]%
        {2h}
\bibfield{author}{\bibinfo{person}{Steven Morad}, \bibinfo{person}{Ryan Kortvelesy}, \bibinfo{person}{Matteo Bettini}, \bibinfo{person}{Stephan Liwicki}, {and} \bibinfo{person}{Amanda Prorok}.} \bibinfo{year}{2023}\natexlab{}.
\newblock \showarticletitle{{POPGym}: Benchmarking partially observable reinforcement learning}.
\newblock \bibinfo{journal}{\emph{arXiv preprint arXiv:2303.01859}} (\bibinfo{year}{2023}).
\newblock


\bibitem[\protect\citeauthoryear{Ono, Fuchs, Steffy, Maimone, and Yen}{Ono et~al\mbox{.}}{2015}]%
        {3o}
\bibfield{author}{\bibinfo{person}{Masahiro Ono}, \bibinfo{person}{Thoams~J Fuchs}, \bibinfo{person}{Amanda Steffy}, \bibinfo{person}{Mark Maimone}, {and} \bibinfo{person}{Jeng Yen}.} \bibinfo{year}{2015}\natexlab{}.
\newblock \showarticletitle{Risk-aware planetary rover operation: Autonomous terrain classification and path planning}. In \bibinfo{booktitle}{\emph{IEEE Aerospace Conference}}.
\newblock


\bibitem[\protect\citeauthoryear{Padalkar, Pooley, Jain, Bewley, Herzog, Irpan, Khazatsky, Rai, Singh, Brohan, et~al\mbox{.}}{Padalkar et~al\mbox{.}}{2023}]%
        {3v}
\bibfield{author}{\bibinfo{person}{Abhishek Padalkar}, \bibinfo{person}{Acorn Pooley}, \bibinfo{person}{Ajinkya Jain}, \bibinfo{person}{Alex Bewley}, \bibinfo{person}{Alex Herzog}, \bibinfo{person}{Alex Irpan}, \bibinfo{person}{Alexander Khazatsky}, \bibinfo{person}{Anant Rai}, \bibinfo{person}{Anikait Singh}, \bibinfo{person}{Anthony Brohan}, {et~al\mbox{.}}} \bibinfo{year}{2023}\natexlab{}.
\newblock \showarticletitle{Open {X}-{E}mbodiment: {R}obotic learning datasets and {RT-X} models}.
\newblock \bibinfo{journal}{\emph{arXiv preprint arXiv:2310.08864}} (\bibinfo{year}{2023}).
\newblock


\bibitem[\protect\citeauthoryear{Patle, Pandey, Parhi, Jagadeesh, et~al\mbox{.}}{Patle et~al\mbox{.}}{2019}]%
        {1a}
\bibfield{author}{\bibinfo{person}{BK Patle}, \bibinfo{person}{Anish Pandey}, \bibinfo{person}{DRK Parhi}, \bibinfo{person}{AJDT Jagadeesh}, {et~al\mbox{.}}} \bibinfo{year}{2019}\natexlab{}.
\newblock \showarticletitle{A review: {O}n path planning strategies for navigation of mobile robot}.
\newblock \bibinfo{journal}{\emph{Defence Technology}} (\bibinfo{year}{2019}).
\newblock


\bibitem[\protect\citeauthoryear{Puthumanaillam, Liu, Mehr, and Ornik}{Puthumanaillam et~al\mbox{.}}{2024a}]%
        {puthumanaillam2024weathering}
\bibfield{author}{\bibinfo{person}{Gokul Puthumanaillam}, \bibinfo{person}{Xiangyu Liu}, \bibinfo{person}{Negar Mehr}, {and} \bibinfo{person}{Melkior Ornik}.} \bibinfo{year}{2024}\natexlab{a}.
\newblock \showarticletitle{Weathering ongoing uncertainty: {L}earning and planning in a time-varying partially observable environment}. In \bibinfo{booktitle}{\emph{2024 IEEE International Conference on Robotics and Automation (ICRA)}}. IEEE, \bibinfo{pages}{4612--4618}.
\newblock


\bibitem[\protect\citeauthoryear{Puthumanaillam, Song, Yesmagambet, Park, and Ornik}{Puthumanaillam et~al\mbox{.}}{2024b}]%
        {puthumanaillam2024tab}
\bibfield{author}{\bibinfo{person}{Gokul Puthumanaillam}, \bibinfo{person}{Jae~Hyuk Song}, \bibinfo{person}{Nurzhan Yesmagambet}, \bibinfo{person}{Shinkyu Park}, {and} \bibinfo{person}{Melkior Ornik}.} \bibinfo{year}{2024}\natexlab{b}.
\newblock \showarticletitle{{TAB-Fields: A} Maximum Entropy Framework for Mission-Aware Adversarial Planning}.
\newblock \bibinfo{journal}{\emph{arXiv preprint arXiv:2412.02570}} (\bibinfo{year}{2024}).
\newblock


\bibitem[\protect\citeauthoryear{Puthumanaillam, Vora, and Ornik}{Puthumanaillam et~al\mbox{.}}{2024c}]%
        {puthumanaillam2024comtraq}
\bibfield{author}{\bibinfo{person}{Gokul Puthumanaillam}, \bibinfo{person}{Manav Vora}, {and} \bibinfo{person}{Melkior Ornik}.} \bibinfo{year}{2024}\natexlab{c}.
\newblock \showarticletitle{{ComTraQ-MPC}: {M}eta-Trained DQN-MPC Integration for Trajectory Tracking with Limited Active Localization Updates}.
\newblock \bibinfo{journal}{\emph{arXiv preprint arXiv:2403.01564}} (\bibinfo{year}{2024}).
\newblock


\bibitem[\protect\citeauthoryear{Rana, Talbot, Dasagi, Milford, and S{\"u}nderhauf}{Rana et~al\mbox{.}}{2020}]%
        {1j}
\bibfield{author}{\bibinfo{person}{Krishan Rana}, \bibinfo{person}{Ben Talbot}, \bibinfo{person}{Vibhavari Dasagi}, \bibinfo{person}{Michael Milford}, {and} \bibinfo{person}{Niko S{\"u}nderhauf}.} \bibinfo{year}{2020}\natexlab{}.
\newblock \showarticletitle{Residual reactive navigation: {C}ombining classical and learned navigation strategies for deployment in unknown environments}. In \bibinfo{booktitle}{\emph{2020 IEEE International Conference on Robotics and Automation}}.
\newblock


\bibitem[\protect\citeauthoryear{Rojas, Padr{\~a}o, Fuentes, Reis, Albayrak, Osmanoglu, and Bobadilla}{Rojas et~al\mbox{.}}{2024}]%
        {w5}
\bibfield{author}{\bibinfo{person}{Cesar~A Rojas}, \bibinfo{person}{Paulo Padr{\~a}o}, \bibinfo{person}{Jose Fuentes}, \bibinfo{person}{Gregory~M Reis}, \bibinfo{person}{Arif~R Albayrak}, \bibinfo{person}{Batuhan Osmanoglu}, {and} \bibinfo{person}{Leonardo Bobadilla}.} \bibinfo{year}{2024}\natexlab{}.
\newblock \showarticletitle{Combining multi-satellite remote and in-situ sensing for unmanned underwater vehicle state estimation}.
\newblock \bibinfo{journal}{\emph{Ocean Engineering}} (\bibinfo{year}{2024}).
\newblock


\bibitem[\protect\citeauthoryear{Shah, Osi{\'n}ski, Levine, et~al\mbox{.}}{Shah et~al\mbox{.}}{2023}]%
        {3u}
\bibfield{author}{\bibinfo{person}{Dhruv Shah}, \bibinfo{person}{B{\l}a{\.z}ej Osi{\'n}ski}, \bibinfo{person}{Sergey Levine}, {et~al\mbox{.}}} \bibinfo{year}{2023}\natexlab{}.
\newblock \showarticletitle{{LM-Nav}: Robotic navigation with large pre-trained models of language, vision, and action}. In \bibinfo{booktitle}{\emph{Conference on Robot Learning}}.
\newblock


\bibitem[\protect\citeauthoryear{Singi, He, Pan, Patel, Sigurdsson, Piramuthu, Song, and Ciocarlie}{Singi et~al\mbox{.}}{2024}]%
        {2i}
\bibfield{author}{\bibinfo{person}{Siddharth Singi}, \bibinfo{person}{Zhanpeng He}, \bibinfo{person}{Alvin Pan}, \bibinfo{person}{Sandip Patel}, \bibinfo{person}{Gunnar~A Sigurdsson}, \bibinfo{person}{Robinson Piramuthu}, \bibinfo{person}{Shuran Song}, {and} \bibinfo{person}{Matei Ciocarlie}.} \bibinfo{year}{2024}\natexlab{}.
\newblock \showarticletitle{Decision making for human-in-the-loop robotic agents via uncertainty-aware reinforcement learning}. In \bibinfo{booktitle}{\emph{2024 IEEE International Conference on Robotics and Automation}}.
\newblock


\bibitem[\protect\citeauthoryear{Spaan, Veiga, and Lima}{Spaan et~al\mbox{.}}{2015}]%
        {1m}
\bibfield{author}{\bibinfo{person}{Matthijs~TJ Spaan}, \bibinfo{person}{Tiago~S Veiga}, {and} \bibinfo{person}{Pedro~U Lima}.} \bibinfo{year}{2015}\natexlab{}.
\newblock \showarticletitle{Decision-theoretic planning under uncertainty with information rewards for active cooperative perception}.
\newblock \bibinfo{journal}{\emph{Autonomous Agents and Multi-Agent Systems}} (\bibinfo{year}{2015}).
\newblock


\bibitem[\protect\citeauthoryear{Srinivasan, Eysenbach, Ha, Tan, and Finn}{Srinivasan et~al\mbox{.}}{2020}]%
        {3t}
\bibfield{author}{\bibinfo{person}{Krishnan Srinivasan}, \bibinfo{person}{Benjamin Eysenbach}, \bibinfo{person}{Sehoon Ha}, \bibinfo{person}{Jie Tan}, {and} \bibinfo{person}{Chelsea Finn}.} \bibinfo{year}{2020}\natexlab{}.
\newblock \showarticletitle{Learning to be safe: Deep {RL} with a safety critic}.
\newblock \bibinfo{journal}{\emph{arXiv preprint arXiv:2010.14603}} (\bibinfo{year}{2020}).
\newblock


\bibitem[\protect\citeauthoryear{Taylor, Berrueta, and Murphey}{Taylor et~al\mbox{.}}{2021}]%
        {1l}
\bibfield{author}{\bibinfo{person}{Annalisa~T Taylor}, \bibinfo{person}{Thomas~A Berrueta}, {and} \bibinfo{person}{Todd~D Murphey}.} \bibinfo{year}{2021}\natexlab{}.
\newblock \showarticletitle{Active learning in robotics: {A} review of control principles}.
\newblock \bibinfo{journal}{\emph{Mechatronics}} (\bibinfo{year}{2021}).
\newblock


\bibitem[\protect\citeauthoryear{Thrun}{Thrun}{2002}]%
        {1t}
\bibfield{author}{\bibinfo{person}{Sebastian Thrun}.} \bibinfo{year}{2002}\natexlab{}.
\newblock \showarticletitle{Particle Filters in Robotics}. In \bibinfo{booktitle}{\emph{UAI}}.
\newblock


\bibitem[\protect\citeauthoryear{Wang, Wang, Zhao, Wang, and Li}{Wang et~al\mbox{.}}{2022}]%
        {2f}
\bibfield{author}{\bibinfo{person}{Ning Wang}, \bibinfo{person}{Yabiao Wang}, \bibinfo{person}{Yuming Zhao}, \bibinfo{person}{Yong Wang}, {and} \bibinfo{person}{Zhigang Li}.} \bibinfo{year}{2022}\natexlab{}.
\newblock \showarticletitle{Sim-to-real: {M}apless navigation for USVs using deep reinforcement learning}.
\newblock \bibinfo{journal}{\emph{Journal of Marine Science and Engineering}} (\bibinfo{year}{2022}).
\newblock


\bibitem[\protect\citeauthoryear{Wang, Yan, Wang, Xu, Wu, and Wang}{Wang et~al\mbox{.}}{2024}]%
        {1v}
\bibfield{author}{\bibinfo{person}{Zixiang Wang}, \bibinfo{person}{Hao Yan}, \bibinfo{person}{Zhuoyue Wang}, \bibinfo{person}{Zhengjia Xu}, \bibinfo{person}{Zhizhong Wu}, {and} \bibinfo{person}{Yining Wang}.} \bibinfo{year}{2024}\natexlab{}.
\newblock \showarticletitle{Research on autonomous robots navigation based on reinforcement learning}. In \bibinfo{booktitle}{\emph{International Conference on Robotics, Artificial Intelligence and Intelligent Control}}. IEEE.
\newblock


\bibitem[\protect\citeauthoryear{Yang, Yang, Dyer, He, Smola, and Hovy}{Yang et~al\mbox{.}}{2016}]%
        {g3}
\bibfield{author}{\bibinfo{person}{Zichao Yang}, \bibinfo{person}{Diyi Yang}, \bibinfo{person}{Chris Dyer}, \bibinfo{person}{Xiaodong He}, \bibinfo{person}{Alex Smola}, {and} \bibinfo{person}{Eduard Hovy}.} \bibinfo{year}{2016}\natexlab{}.
\newblock \showarticletitle{Hierarchical attention networks for document classification}. In \bibinfo{booktitle}{\emph{Conference of the North American Chapter of the Association for Computational Linguistics: Human Language Technologies}}.
\newblock


\bibitem[\protect\citeauthoryear{Yarats, Zhang, Kostrikov, Amos, Pineau, and Fergus}{Yarats et~al\mbox{.}}{2021}]%
        {2w}
\bibfield{author}{\bibinfo{person}{Denis Yarats}, \bibinfo{person}{Amy Zhang}, \bibinfo{person}{Ilya Kostrikov}, \bibinfo{person}{Brandon Amos}, \bibinfo{person}{Joelle Pineau}, {and} \bibinfo{person}{Rob Fergus}.} \bibinfo{year}{2021}\natexlab{}.
\newblock \showarticletitle{Improving sample efficiency in model-free reinforcement learning from images}. In \bibinfo{booktitle}{\emph{AAAI Conference on Artificial Intelligence}}.
\newblock


\bibitem[\protect\citeauthoryear{Zhang, Sreedharan, Kulkarni, Chakraborti, Zhuo, and Kambhampati}{Zhang et~al\mbox{.}}{2016}]%
        {3p}
\bibfield{author}{\bibinfo{person}{Yu Zhang}, \bibinfo{person}{Sarath Sreedharan}, \bibinfo{person}{Anagha Kulkarni}, \bibinfo{person}{Tathagata Chakraborti}, \bibinfo{person}{Hankz~Hankui Zhuo}, {and} \bibinfo{person}{Subbarao Kambhampati}.} \bibinfo{year}{2016}\natexlab{}.
\newblock \showarticletitle{Plan explicability for robot task planning}. In \bibinfo{booktitle}{\emph{RSS Workshop on Planning for Human-Robot Interaction: Shared Autonomy and Collaborative Robotics}}.
\newblock


\bibitem[\protect\citeauthoryear{Zhu and Zhang}{Zhu and Zhang}{2021}]%
        {2l}
\bibfield{author}{\bibinfo{person}{Kai Zhu} {and} \bibinfo{person}{Tao Zhang}.} \bibinfo{year}{2021}\natexlab{}.
\newblock \showarticletitle{Deep reinforcement learning based mobile robot navigation: {A} review}.
\newblock \bibinfo{journal}{\emph{Tsinghua Science and Technology}} (\bibinfo{year}{2021}).
\newblock


\bibitem[\protect\citeauthoryear{Zhu, Wang, Chen, and Dong}{Zhu et~al\mbox{.}}{2021}]%
        {1n}
\bibfield{author}{\bibinfo{person}{Yuanyang Zhu}, \bibinfo{person}{Zhi Wang}, \bibinfo{person}{Chunlin Chen}, {and} \bibinfo{person}{Daoyi Dong}.} \bibinfo{year}{2021}\natexlab{}.
\newblock \showarticletitle{Rule-based reinforcement learning for efficient robot navigation with space reduction}.
\newblock \bibinfo{journal}{\emph{IEEE/ASME Transactions on Mechatronics}} (\bibinfo{year}{2021}).
\newblock


\end{thebibliography}


\end{document}